\PassOptionsToPackage{table,xcdraw}{xcolor}
\documentclass[11pt]{article}

\usepackage[final]{acl}

\usepackage{times}
\usepackage{latexsym}

\usepackage[T1]{fontenc}

\usepackage[utf8]{inputenc}

\usepackage{microtype}

\usepackage{inconsolata}

\usepackage{graphicx}

\usepackage{multirow}
\usepackage[table,xcdraw]{xcolor}
\usepackage{amsmath}
\usepackage{booktabs}
\usepackage{subcaption}
\usepackage{comment}
\usepackage{amssymb}
\usepackage{url}
\title{\mbox{OSCBench: Benchmarking Object State Change in Text-to-Video Generation}}

\author{
 \textbf{Xianjing Han \textsuperscript{1} \textsuperscript{*}},
 \textbf{Bin Zhu\textsuperscript{2} \textsuperscript{*} \textsuperscript{\dag}},
 \textbf{Shiqi Hu\textsuperscript{1}},
 \textbf{Franklin Mingzhe Li\textsuperscript{3}},
\\
 \textbf{Patrick Carrington\textsuperscript{3}},
 \textbf{Roger Zimmermann\textsuperscript{1}},
 \textbf{Jingjing Chen\textsuperscript{4}}
\\
\\
 \textsuperscript{1} National University of Singapore
 \textsuperscript{2} Singapore Management University\\
 \textsuperscript{3} Carnegie Mellon University
 \textsuperscript{4} Fudan University
\\
 \small{
   Correspondence: \href{mailto:binzhu@smu.edu.sg}{binzhu@smu.edu.sg}
 }
}

\begin{document}
\maketitle
\footnotetext[1]{\textsuperscript{*} Equal contribution.}
\footnotetext[2]{\textsuperscript{\textdagger} Corresponding author and project lead.}

\begin{abstract}
Text-to-video (T2V) generation models have made rapid progress in producing visually high-quality and temporally coherent videos. However, existing benchmarks primarily focus on perceptual quality, text–video alignment, or physical plausibility, leaving a critical aspect of action understanding largely unexplored: object state change (OSC) explicitly specified in the text prompt. OSC refers to the transformation of an object’s state induced by an action, such as peeling a potato or slicing a lemon. 
In this paper, we introduce OSCBench, a benchmark specifically designed to assess OSC performance in T2V models. OSCBench is constructed from instructional cooking data and systematically organizes action–object interactions into regular, novel, and compositional scenarios to probe both in-distribution performance and generalization.
We evaluate six representative open-source and proprietary T2V models using both human user study and multimodal large language model (MLLM)–based automatic evaluation. Our results show that, despite strong performance on semantic and scene alignment, current T2V models consistently struggle with accurate and temporally consistent object state changes, especially in novel and compositional settings. These findings position OSC as a key bottleneck in text-to-video generation and establish OSCBench as a diagnostic benchmark for advancing state-aware video generation models.
Project page: \url{https://hanxjing.github.io/OSCBench}.
\end{abstract}

\begin{figure*}[t]
	\centering
	\includegraphics[width=0.97\textwidth]{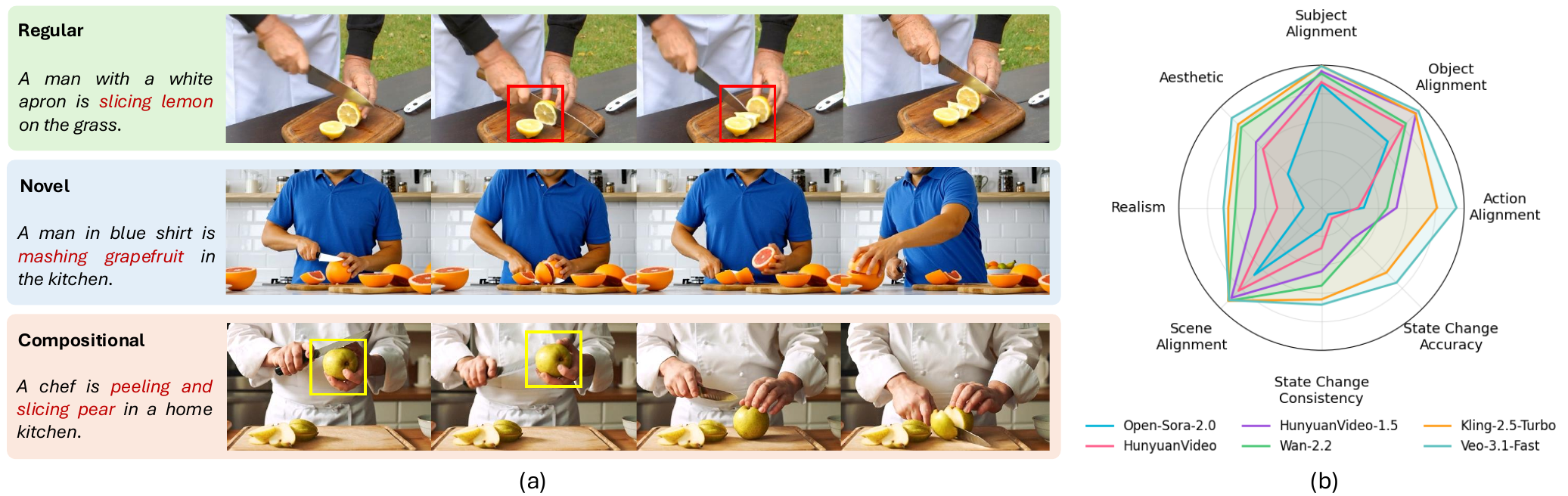}
    \vspace{-1mm}
	\caption{Overview of OSCBench evaluation. (a) Representative failure cases from regular, novel, and compositional object state change scenarios. In the regular case, the red box marks an implausible state change of the lemon during slicing. In the novel case, the model misinterprets the instructed action, resulting in a wrong object transformation. In the compositional case, the yellow box indicates an incomplete state change where the pear remains unpeeled. (b) Human-evaluated multi-dimensional performance of T2V models on OSCBench.}
	\label{intro}
\end{figure*}

\section{Introduction}
Text-to-video (T2V) generation models have made remarkable progress in recent years, producing videos with increasingly high visual fidelity and temporal coherence. These advances have enabled a wide range of applications, including creative content generation, instructional video synthesis, and simulation of real-world processes~\citep{Veo2025,ma2025step}. 
As T2V models continue to scale, a central question emerges: to what extent do these models faithfully realize the consequences of actions specified in language~\cite{pearl2009causality, zhu2020cookgan, souvcek2024genhowto}, rather than merely producing visually appealing motion patterns?

Recent benchmarks have taken important steps toward answering this question by evaluating physical plausibility and commonsense constraints in generated videos, such as adherence to gravity, collisions, and material properties~\citep{meng2024towards,gu2025phyworldbench}. While these evaluations probe fundamental aspects of physical realism, they overlook a critical dimension of language-grounded action understanding that is ubiquitous in everyday activities: Object State Change (OSC) explicitly specified by the prompt. In many real-world tasks, such as slicing a lemon, peeling a carrot, or mixing dough, success is defined not only by performing an action, but by transforming an object from an initial state to a specific target state (e.g., a whole lemon becoming sliced)~\cite{souvcek2022look, xue2024learning, li2025oscar}. Correctly modeling such object state change is essential for downstream applications, including robotics, embodied AI, and instructional video generation. 

Object state change poses a particularly stringent test of language-grounded reasoning in T2V models. Correct OSC generation requires a model to understand the action semantics expressed in language, infer the intended object transformation, and render a continuous and coherent visual evolution over time. However, despite producing visually compelling videos, current T2V models often fail on this dimension: generated outputs may appear realistic at a glance while exhibiting incorrect, incomplete, or inconsistent object state changes.
Figure~\ref{intro} (a) illustrates representative failure cases, where objects change into implausible states or the instructed action is misunderstood, revealing a gap between high-level semantic alignment and faithful realization of action consequences.
Despite the importance of OSC, it has not been systematically evaluated in existing T2V benchmarks, which primarily emphasize overall perceptual quality, text–video alignment, or physical plausibility, without explicitly assessing whether an object reaches the correct target state or whether the state transition unfolds consistently over time.

To address this gap, we introduce OSCBench, a benchmark designed to evaluate object state change in text-to-video generation. We focus on instructional cooking scenarios, where state changes are frequent, diverse, and well-defined, and build OSCBench on top of the HowToChange dataset~\citep{xue2024learning}.
To enable balanced and comprehensive evaluation, we abstract actions and objects into semantically meaningful categories and then construct three complementary evaluation regimes as shown in Figure~\ref{intro} (a): regular scenarios covering common action–object pairs (e.g., slicing lemon), novel scenarios that test generalization to uncommon yet feasible state changes (e.g., mashing grapefruit), and compositional scenarios involving multiple action compositions (e.g., peeling and slicing pear).
In total, OSCBench comprises 1,120 prompts across 140 object-state scenarios, providing a specific benchmark for evaluating OSC performance in T2V models. 

In addition, we evaluate six state-of-the-art (SOTA) T2V models on OSCBench, including four widely used open-source systems (Open-Sora-2.0~\citep{opensora2}, HunyuanVideo~\citep{kong2024hunyuanvideo}, HunyuanVideo-1.5~\citep{hunyuanvideo2025}, Wan-2.2~\citep{wan2025}) and two proprietary models (Kling-2.5-Turbo~\citep{KlingAI2024} and Veo-3.1-Fast~\citep{Veo2025}). We conduct both human user study and automatic evaluation using the latest multimodal large language models (MLLMs).
Across the two evaluation methods, we design a comprehensive set of criteria covering semantic adherence, OSC performance, scene alignment, and perceptual quality. In particular, rather than using MLLMs as black-box scorers, we employ Chain-of-Thought~\citep{wei2022chain} evaluation strategy that explicitly guides the reasoning process through criteria grounding, evidence extraction, and score justification. We further analyze the correlation between human judgments and MLLM-based evaluations to assess the reliability of automated OSC evaluation.
Our results in Figure~\ref{intro} (b) reveal that while SOTA T2V models generally perform well on high-level semantic alignment (e.g., subject, object, and scene), object state change accuracy and consistency remain a significant challenge.
These findings position OSC as a critical diagnostic dimension that complements existing evaluations. By revealing how state changes deviate from intended action effects, OSCBench provides practical guidance for building video generation models that reason more faithfully about actions and their consequences.

In summary, our contributions are three-fold:
\begin{itemize}
\setlength{\itemsep}{0pt}
\setlength{\parskip}{0pt}
\item We introduce OSCBench, the first benchmark explicitly designed to evaluate object state change in text-to-video generation across regular, novel, and compositional scenarios.
\item We design a set of criteria covering semantic adherence, OSC performance, scene alignment, and perceptual quality to comprehensively evaluate the video generation performance with both human user study and automatic MLLM assessment.
\item We benchmark six SOTA T2V models, systematically examine their performance across different OSC scenarios, and identify key challenges that persist. The results offer guidance for designing models with OSC-aware generation and outline directions for future research.
\end{itemize}

\section{Related Work}
\textbf{Benchmarks for Text-to-Video Generation.}
The rapid advancement of T2V models has motivated the development of benchmarks for accurate and reliable assessment. A number of recent benchmarks~\citep{huang2024vbench,he2024videoscore} aim to provide systematic evaluation of T2V models either from a comprehensive perspective or through specific aspects of generation quality. For example, VBench~\citep{huang2024vbench} and Eval-Crafter~\citep{liu2024evalcrafter} target holistic evaluation across multiple interpretable dimensions, including temporal consistency, motion smoothness, and text–video alignment.
To better diagnose particular modeling challenges, several aspect-specific benchmarks have been proposed. For example, T2V-CompBench~\citep{sun2025t2v} evaluates compositional generation capabilities, while DEVIL~\citep{liao2024evaluation} focuses on the dynamic characteristics of generated videos. More recently, researchers have observed that T2V models frequently generate videos that violate physical constraints. This has motivated the development of benchmarks that explicitly assess physical plausibility, such as VideoPhy~\citep{bansal2024videophy}, PhyGenBench~\citep{meng2024towards}, and PhyWorldBench~\citep{gu2025phyworldbench}, which examine whether generated videos adhere to basic physical commonsense.
Despite these advances, existing benchmarks pay limited attention to OSC. In this work, we introduce a benchmark specifically for object state change, providing scenarios that require accurate state modeling and enabling targeted evaluation of a model’s OSC understanding.

\noindent\textbf{Evaluation Methods for Text-to-Video Models.}
Recent video benchmarks~\citep{huang2024vbench,meng2024towards,gu2025phyworldbench} commonly adopt a hybrid evaluation protocol that combines automatic model evaluation with human user study. For automatic evaluation, CLIP~\citep{xue2024learning} and ViCLIP~\citep{wang2023internvid} based text–video similarity models are widely used to assess semantic alignment between prompts and generated videos.
More recently, MLLMs have demonstrated strong abilities in understanding complex visual content~\citep{ouyang2025punchbench,zhang-etal-2025-redundancy-principles,he2025mmboundary}. Therefore, many video benchmarks~\citep{feng2025tc,motamed2025generative,han2025video} employ MLLMs to evaluate the semantic consistency in generated videos. Building on this capability, PhyWorldBench~\citep{gu2025phyworldbench} further leverages MLLMs to evaluate whether generated videos obey physical laws, which often requires multi-step reasoning.
To evaluate fine-grained OSC, we leverage the reasoning capabilities of MLLMs and adopt a CoT strategy~\citep{wei2022chain}. Unlike existing benchmarks~\citep{gu2025phyworldbench}, which mainly use CoT to generate textual descriptions, we use it to guide models through a structured reasoning process, encouraging careful visual inspection and more reliable state-change judgments.

\begin{figure*}[t]
	\centering
	\includegraphics[width=\textwidth]{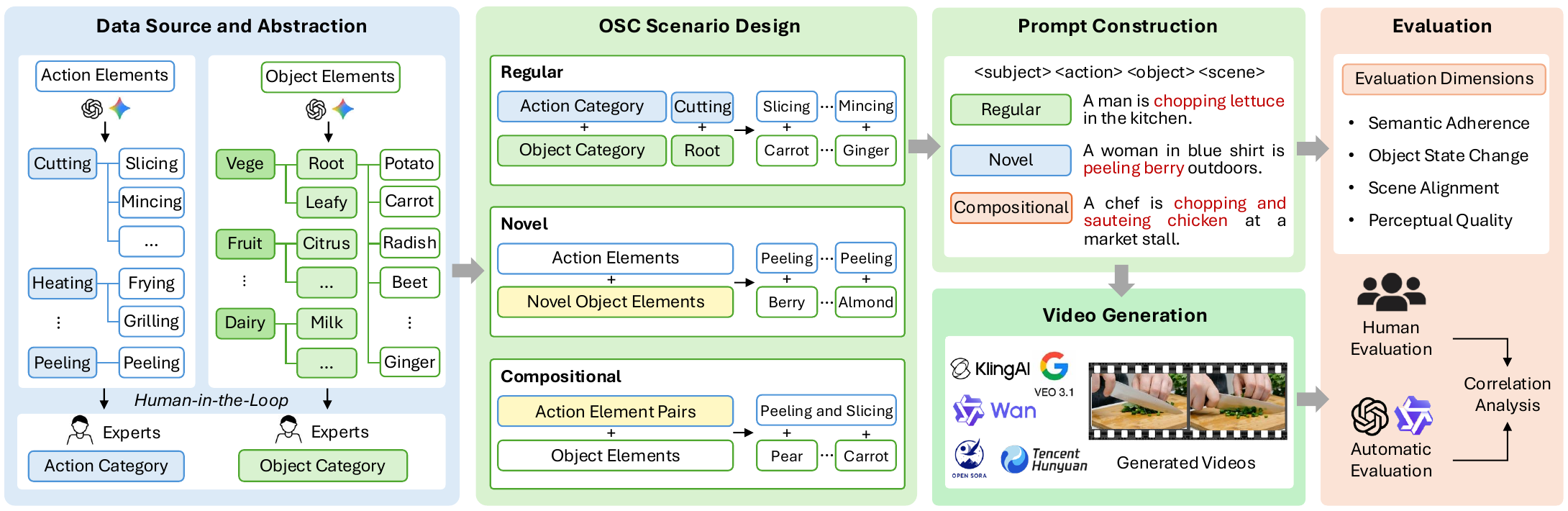}
	\caption{Overview of the OSCBench construction and evaluation pipeline. We build unified action and object categories from instructional cooking data via a human-in-the-loop process, and construct regular, novel, and compositional OSC scenarios as text prompts for video generation. The generated videos are evaluated by humans and MLLMs across multiple criteria, and we analyze their correlations to assess automatic evaluation reliability.}
	\label{framework}
\end{figure*}

\section{OSCBench Construction}
The goal of OSCBench is to provide a structured and comprehensive benchmark for evaluating object state change in text-to-video generation. Designing such a benchmark requires addressing three key challenges: (i) covering realistic and diverse object state changes grounded in textual prompts, (ii) ensuring controlled and balanced coverage of actions and objects to reduce dataset bias, and (iii) introducing varying levels of difficulty to probe both memorization and generalization. In this section, we describe how OSCBench is constructed to meet these requirements.

\subsection{Data Source and Abstraction}
Object state change is ubiquitous in everyday activities, with cooking being a representative domain. Cooking tasks naturally involve diverse state transformations, such as chopping, peeling, and heating, and exhibit clear causal relationships between actions and resulting object states. We therefore build OSCBench on the HowToChange dataset~\citep{xue2024learning}, which is derived from instructional cooking videos in HowTo100M~\citep{miech2019howto100m}. HowToChange contains 20 fine-grained action elements and 134 object elements, yielding 409 distinct action–object combinations (e.g., slicing apple).
However, these combinations exhibit a strong long-tail distribution: common pairs appear frequently (e.g., chopping potato), while many plausible ones are rare or absent (e.g., squeezing ginger). Directly sampling from this distribution would bias evaluation toward frequent patterns and limit insights into generalization.

To mitigate this issue, we reorganize the raw action and object elements into high-level conceptual categories using a human-in-the-loop abstraction process. Specifically, as shown in Figure~\ref{framework}, guided by cooking objectives, we first use GPT-5.2~\citep{OpenAI_GPT52_2025} and Gemini-3~\citep{Google_Gemini3_2025} to propose candidate groupings of the 20 action elements into 9 action categories (e.g., heating), and to cluster the 134 object elements into 8 major object categories (e.g., vegetable) with 28 finer-grained subcategories (e.g., root vegetables). These groupings are then iteratively refined and validated by human experts to ensure semantic correctness and practical plausibility. This abstraction enables systematic scenario construction while preserving semantic diversity.

\subsection{OSC Scenario Design}
Based on the abstracted action and object taxonomy, we construct three complementary types of OSC scenarios to evaluate different aspects of OSC-aware video generation: regularity, generalization, and compositionality.

\noindent \textbf{Regular OSC Scenarios}.
Regular scenarios are designed to cover a broad range of realistic and commonly occurring object state changes. We pair each action category with compatible object subcategories to form candidate scenarios. All candidates are first filtered using automated checks by ChatGPT and then validated by human review. This process yields 108 regular OSC scenarios. For each scenario, we further enumerate concrete instances by pairing specific action elements with object elements and manually select 8 representative action–object combinations (e.g., mincing ginger), ensuring diversity while maintaining feasibility.

\noindent \textbf{Novel OSC Scenarios}.
To evaluate whether models can reason about unfamiliar yet plausible object state changes, we introduce novel scenarios that deliberately deviate from common action-object combinations. 
For each of the 20 action elements, we select 8 uncommon yet physically feasible objects that were verified by human inspection (e.g., peeling berries), resulting in 20 novel scenarios. These scenarios cannot be reliably solved through memorization of frequent action–object pairs and instead require models to infer state changes from action semantics.

\noindent \textbf{Compositional OSC Scenarios}.
Real-world activities often involve multiple actions applied sequentially, where state changes evolve over time. To assess whether models can maintain coherent intermediate and final states, we construct compositional scenarios by composing pairs of action elements (e.g., peeling followed by slicing). We select 12 common action pairs, verified by human inspection, and combine each pair with 8 suitable objects (e.g., peeling and slicing potato). These scenarios explicitly examine multiple action composition and temporal consistency for OSC-aware video generation.

\subsection{Prompt Construction}
For every action–object combination in each scenario, we generate prompts using a structured template: <subject><action><object><scene>.
We randomly generate three candidate prompts for each combination using GPT-5.2 and manually select the most natural one. Examples can be seen in Figure~\ref{framework} (e.g., A man is chopping lettuce in the kitchen). In addition to full prompts with subjects and scenes, we further test how models respond when only object state change cues are provided. Specifically, we randomly simplify 1-2 prompts per scenario to the minimal form, <action><object>. This variant reduces contextual cues and places greater emphasis on the model’s ability to infer and realize OSC directly from the action description.

\subsection{Benchmark Statistics}
OSCBench comprises 140 object state change scenarios in total, including 108 regular scenarios, 20 novel scenarios, and 12 compositional scenarios. Each scenario contains 8 action–object combinations, resulting in 1,120 prompts overall. The prompts are concise and descriptive, with an average length of 9.2 words, providing sufficient context while avoiding unnecessary linguistic complexity. We additionally provide a word cloud for OSCBench to illustrate the word distribution in the Appendix~\ref{OSCBench_detail}. By combining structured abstraction, controlled scenario design, and multiple difficulty regimes, OSCBench enables systematic analysis of object state change performance in text-to-video models, covering both common patterns and challenging generalization cases.

\section{Evaluation}
Evaluating T2V generation models is inherently challenging, particularly when the goal is to assess object state change specified by the prompt. A reliable evaluation must verify not only whether a generated video aligns with the prompt at a semantic level, but also whether the prompt-implied object state transition is realized accurately and consistently over time.
While human evaluators can naturally perform such judgments, large-scale human evaluation is costly and difficult to scale. Following PhyWorldBench~\citep{gu2025phyworldbench}, we conduct both human user study and automatic assessment using multiple large language models.

\subsection{Evaluation Dimensions}
\label{sec:evaluation_criteria}
We comprehensively evaluate generated videos along four complementary evaluation dimensions: semantic adherence, object state change, scene alignment, and perceptual quality.

\noindent \textbf{Semantic Adherence}.
This dimension measures whether the core semantic entities described in the prompt are faithfully grounded in the generated video. Specifically, we evaluate three key components independently: Subject alignment to measure whether the acting subject (e.g., a man or a woman) is present and correct, object alignment to evaluate whether the manipulated object matches the prompt and action alignment to assess whether the performed action corresponds to the intended action described in the prompt.

\noindent \textbf{Object State Change}. This is the central dimension of OSCBench. Evaluating object state change requires reasoning about both the outcome and the temporal evolution of the object. We therefore decompose OSC evaluation into two sub-dimensions: state-change accuracy, which measures whether the object reaches the correct target state implied by the prompt (e.g., a whole apple becoming sliced), and state-change consistency, which assesses whether the transformation unfolds smoothly and coherently over time, without abrupt jumps or unnatural object appearances or unexplained appearance or disappearance of object parts.

\noindent \textbf{Scene Alignment}.
This dimension evaluates whether the global environment in the video matches the scene description in the prompt (e.g., kitchen or market). It focuses on the background context, such as whether the video clearly occurs in a kitchen or an outdoor market, and whether the scene remains stable and coherent over time. 

\noindent \textbf{Perceptual Quality}.
This dimension measures the overall visual impression of the video and includes two aspects: realism, which measures whether the video resembles real-world footage in motion, lighting, and texture, and aesthetic quality, which reflects how visually appealing the video appears in composition, color, and overall presentation.

\subsection{Human Evaluation}
We first conduct human user study as a strong reference to evaluate our OSCBench. As exhaustive human evaluation over all generated videos would be prohibitively costly and time-consuming, we adopt a representative sampling strategy. Specifically, to cover the full diversity of OSCBench, we sample one prompt from each of the 140 OSC scenarios, ensuring that all regular, novel, and compositional scenarios are represented. For each selected prompt, we generate one video for each T2V model, resulting in 140 videos per model for human evaluation.
Each video is independently rated by three human evaluators across the evaluation dimensions described in Section~\ref{sec:evaluation_criteria}.
To encourage fine-grained and consistent judgments, we provide a 1-5 Likert scale for each dimension. For each text–video pair, we average the three evaluators' scores to obtain the mean opinion score for each evaluation dimension. These human scores serve both as primary benchmark results and as a reference signal for validating automatic evaluation using MLLMs.

\begin{table*}[t]
	\centering
	\scriptsize
	{
		\renewcommand{\arraystretch}{1.02}
		\begin{tabular}{p{5cm}cccccccc}
			\hline
			\multicolumn{1}{c}{} & \multicolumn{3}{c}{Semantic Adherence}        & \multicolumn{2}{c}{Object State Change} & \multicolumn{1}{c}{\multirow{3}{*}{\shortstack{Scene\\Alignment}}}   & \multicolumn{2}{c}{Perceptual Quality} \\ 
			\cmidrule(lr){2-4} \cmidrule(lr){5-6} \cmidrule(lr){8-9}
			\multicolumn{1}{c}{\multirow{-2}{*}{Model}} & \multicolumn{1}{c}{Subject} & \multicolumn{1}{c}{Object} & \multicolumn{1}{c}{Action} & \multicolumn{1}{c}{Accuracy} & \multicolumn{1}{c}{Consistency} & & \multicolumn{1}{c}{Realism} & \multicolumn{1}{c}{Aesthetics} \\ \hline
			\multicolumn{9}{l}{\cellcolor[HTML]{EFEFEF}\textit{Open-source models}}                 \\
			Open-Sora-2.0~\citep{opensora2}   & 0.860& 0.734 & 0.518 & 0.380 & 0.428 & 0.740 & 0.416& 0.540\\
			HunyuanVideo~\citep{kong2024hunyuanvideo}   & 0.868 & 0.826 & 0.494& 0.402 & 0.510  & 0.834  & 0.526& 0.688 \\
			HunyuanVideo-1.5~\citep{hunyuanvideo2025}& \textbf{0.914}& \textbf{0.902}& \textbf{0.656}& 0.524 & 0.608& 0.876& 0.618& 0.730 \\
			Wan-2.2~\citep{wan2025} &0.904& 0.842 & 0.616& \textbf{0.560} & \textbf{0.668}& \textbf{0.894}& \textbf{0.702}&\textbf{0.818}\\ \hline
			\multicolumn{9}{l}{\cellcolor[HTML]{EFEFEF}\textit{Proprietary models}} \\
			Kling-2.5-Turbo~\citep{KlingAI2024}& \textbf{0.938}& 0.900& 0.826& 0.726& 0.726& \textbf{0.894}& 0.732& 0.836 \\
			Veo-3.1-Fast~\citep{Veo2025} & 0.936& \textbf{0.916}& \textbf{0.908}& \textbf{0.786} & \textbf{0.748}& 0.890& \textbf{0.752}& \textbf{0.874}\\ \hline
		\end{tabular}
	}
    \vspace{-0.2cm}
	\caption{Human evaluation results of different T2V models across multiple evaluation dimensions.}
	\label{human}
\end{table*}

\begin{table*}[t]
	\centering
	\scriptsize
	{
		\renewcommand{\arraystretch}{1.02}
		\begin{tabular}{p{5cm}cccccccc}
			\hline
			\multicolumn{1}{c}{} & \multicolumn{3}{c}{Semantic Adherence}        & \multicolumn{2}{c}{Object State Change} & \multicolumn{1}{c}{\multirow{3}{*}{\shortstack{Scene\\Alignment}}}   & \multicolumn{2}{c}{Perceptual Quality} \\ 
			\cmidrule(lr){2-4} \cmidrule(lr){5-6} \cmidrule(lr){8-9}
			\multicolumn{1}{c}{\multirow{-2}{*}{Model}} & \multicolumn{1}{c}{Subject} & \multicolumn{1}{c}{Object} & \multicolumn{1}{c}{Action} & \multicolumn{1}{c}{Accuracy} & \multicolumn{1}{c}{Consistency} & & \multicolumn{1}{c}{Realism} & \multicolumn{1}{c}{Aesthetics} \\ \hline
			\multicolumn{9}{l}{\cellcolor[HTML]{EFEFEF}\textit{Open-source models}}                 \\
			Open-Sora-2.0~\citep{opensora2}   & 0.910& 0.722 & 0.616 & 0.512 & 0.658 & 0.892 & 0.634& 0.712\\
			HunyuanVideo~\citep{kong2024hunyuanvideo}   & 0.898 & 0.764 & 0.562& 0.466 & 0.730  & 0.948  & 0.752& 0.782 \\
			HunyuanVideo-1.5~\citep{hunyuanvideo2025}& \textbf{0.982}&\textbf{0.788}& \textbf{0.642}& \textbf{0.546} & 0.708& 0.936& 0.736& 0.778 \\
			Wan-2.2~\citep{wan2025} &0.950& 0.774 & 0.570 & 0.518 & \textbf{0.710} & \textbf{0.974} &\textbf{0.768} & \textbf{0.798}\\ \hline
			\multicolumn{9}{l}{\cellcolor[HTML]{EFEFEF}\textit{Proprietary models}} \\
			Kling-2.5-Turbo~\citep{KlingAI2024}& \textbf{0.990}& 0.792& 0.742& 0.652& 0.692& 0.972& 0.772& \textbf{0.802} \\
			Veo-3.1-Fast~\citep{Veo2025} & 0.976& \textbf{0.834}& \textbf{0.802}& \textbf{0.740} & \textbf{0.702}& \textbf{0.978}& \textbf{0.782}& \textbf{0.802}\\ \hline
		\end{tabular}
	}
    \vspace{-0.2cm}
	\caption{GPT-5.2–based evaluation results of T2V models on OSCBench across multiple evaluation dimensions.}
	\label{gpt_table}
\end{table*}

\subsection{MLLM-Based Automatic Evaluation}
Automatic evaluation using text–video similarity models (e.g., CLIP and ViCLIP) measures coarse semantic alignment but is insufficient for assessing fine-grained object state changes and perceptual quality.
MLLMs have recently shown strong visual understanding and multi-step reasoning abilities, which can serve as reasoning-based evaluators for video generation~\citep{gu2025phyworldbench}. 
Rather than treating MLLMs as black-box scorers, we design a CoT evaluation strategy that structures the reasoning process. For each video and each evaluation dimension, the MLLM follows three steps:
(1) \textbf{Criteria grounding.} The model restates the scoring criterion of each evaluation dimension in its own words, ensuring it internalizes the scoring definition before examining the video.
(2) \textbf{Evidence extraction.} The model then identifies frame-level visual evidence that is relevant to the criterion and briefly explains why these observations support its assessment.
(3) \textbf{Score decision.} Based on the extracted evidence, the model assigns a discrete score from 1 to 5 and explicitly links the score to the observed evidence. We provide the detailed prompt used for MLLM evaluation in the Appendix~\ref{evaluation_detail}.

We apply this procedure to all adopted MLLMs across all evaluation dimensions. By constraining the reasoning route, the CoT strategy encourages the model to focus on fine-grained object states and their temporal evolution, rather than being distracted by salient but irrelevant visual details.

\section{Evaluation Results and Analysis}
\subsection{Experimental Setup}
We evaluate six representative SOTA T2V generation models, including four widely used open-source systems (Open-Sora-2.0~\citep{opensora2}, HunyuanVideo~\citep{kong2024hunyuanvideo}, HunyuanVideo-1.5~\citep{hunyuanvideo2025}, and Wan-2.2~\citep{wan2025}) and two proprietary models (Kling-2.5-Turbo~\citep{KlingAI2024} and Veo-3.1-Fast~\citep{Veo2025}). Detailed video generation settings are provided in the Appendix~\ref{generation_detail}.
For automatic evaluation, we assess the generated videos using ViCLIP for semantic similarity measurement as well as MLLM, including Qwen3-VL-30B, GPT-5-mini, and GPT-5.2. For space considerations, we present GPT-5.2–based evaluation results in the main paper and include results from other MLLMs in the Appendix~\ref{evaluation_detail}.
All human and automatic evaluation scores are normalized to 0–1 for comparison.

\begin{figure}[t]
	\centering
	\includegraphics[width=0.45\textwidth]{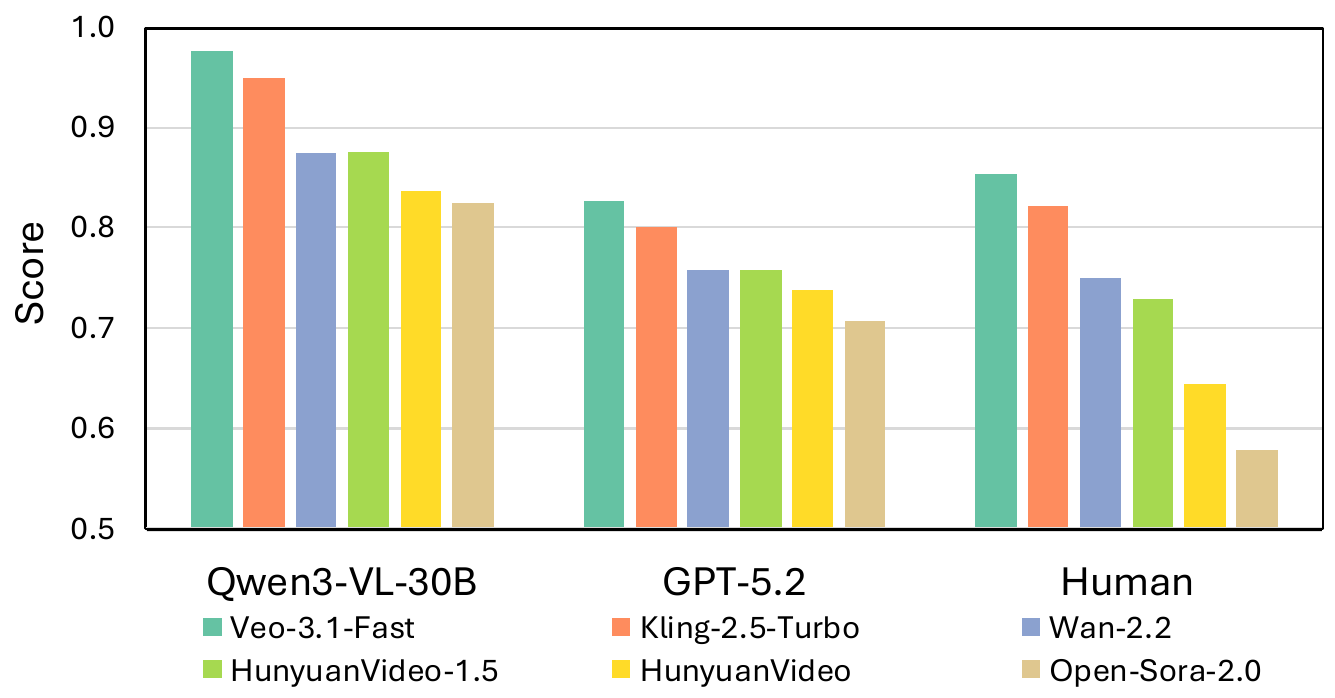}
    \vspace{-0.1cm}
	\caption{Overall performance comparison of T2V models based on aggregated evaluation scores from human evaluator and MLLM-based evaluators (Qwen3-VL-30B and GPT-5.2).}
	\label{bar}
\end{figure}

\begin{table*}[t]
	\centering
	\scriptsize
	{
	\renewcommand{\arraystretch}{1.02}
	\begin{tabular}{p{1.9cm}>{\centering}p{0.43cm}>{\centering}p{0.43cm}>{\centering}p{0.43cm}>{\centering}p{0.43cm}>{\centering}p{0.43cm}>{\centering}p{0.43cm}>{\centering}p{0.4cm}>{\centering}p{0.43cm}>{\centering}p{0.43cm}>{\centering}p{0.43cm}>{\centering}p{0.43cm}>{\centering}p{0.43cm}>{\centering}p{0.43cm}>{\centering}p{0.43cm}>{\centering}p{0.43cm}c}
		\hline
		\multicolumn{1}{c}{\multirow{3}{*}{Metrics}} & \multicolumn{6}{c}{Semantic Adherence} & \multicolumn{4}{c}{Object State Change}    & \multicolumn{2}{c}{\multirow{3}{*}{\shortstack{Scene\\Alignment}}}& \multicolumn{4}{c}{Perceptual Quality}\\ 
		\cmidrule(lr){2-7} \cmidrule(lr){8-11} \cmidrule(lr){14-17}
		\multicolumn{1}{c}{}     & \multicolumn{2}{c}{Subject} & \multicolumn{2}{c}{Object} & \multicolumn{2}{c}{Action} & \multicolumn{2}{c}{Accuracy} & \multicolumn{2}{c}{Consistency} & && \multicolumn{2}{c}{Realism} & \multicolumn{2}{c}{Aesthetics} \\ \cline{2-17} 
		& $\tau$ & $\rho$ & $\tau$ & $\rho$ & $\tau$ & $\rho$ &$\tau$ & $\rho$ & $\tau$ & $\rho$ & $\tau$ & $\rho$ &$\tau$ & $\rho$ & $\tau$ & $\rho$ \\ \hline
		ViCLIP   &0.106&0.132&0.195&0.245&0.288&0.386&-&-&-&-&-&-&-&-&-& \\
		Qwen3-VL-30B   & 0.406  & 0.413  & 0.412  & 0.429 & 0.542 & 0.624 & 0.426&0.503&0.289  & 0.341& 0.145& 0.149& 0.269& 0.297  & 0.407  & 0.426\\
		GPT-5-mini     &\textbf{0.433}&\textbf{0.439}&0.428&0.441&0.478&0.543&0.342&0.392&0.243&0.259&0.200&0.206&0.303&0.338&\textbf{0.514}&\textbf{0.541}  \\
		GPT-5.2 (w/o CoT) & 0.295  & 0.318  & 0.409 & 0.444 & 0.623  & 0.703 & 0.415   & 0.493  &0.303&0.343& 0.425  & 0.447& \textbf{0.323}  & \textbf{0.355}  &0.393&0.411 \\
		GPT-5.2  &0.369 &0.374& \textbf{0.433}  & \textbf{0.466} & \textbf{0.628}  & \textbf{0.710} & \textbf{0.427}   & \textbf{0.507}  & \textbf{0.317} &\textbf{0.359} & \textbf{0.485}  & \textbf{0.505}& 0.276  & 0.318   & 0.367& 0.385   \\ \hline
		Human& 0.468  & 0.472  & 0.484  & 0.506 & 0.636  & 0.735 & 0.603  & 0.691  & 0.501& 0.598& 0.492  & 0.517& 0.613  & 0.711 &0.581& 0.647   \\ \hline
	\end{tabular}
	}
    \vspace{-0.2cm}
    \caption{Correlation between human and MLLM-based automatic evaluations in terms of Kendall's $\tau$ and Spearman's $\rho$. The last row reports the mean inter-human correlation for reference.}
	 \label{correlation}
\end{table*}

\subsection{Performance Comparison}
Figure~\ref{bar} presents the overall evaluation results by averaging the scores of all evaluation dimensions based on Qwen3-VL-30B, GPT-5.2 and human judgment. Among the evaluated models, Veo-3.1-Fast achieves the strongest overall performance, followed by Kling-2.5-Turbo, while open-source models exhibit comparatively lower performance on average.
Tables~\ref{human} and~\ref{gpt_table} report the human evaluation and GPT-5.2-based automatic assessment scores for each model across individual evaluation dimensions, respectively. Despite differences in absolute scores, both evaluation methods exhibit highly consistent trends across models. It can be observed that most models perform well on semantic adherence (particularly for subject and object) and scene alignment, but exhibit substantially lower scores on OSC accuracy and consistency. This discrepancy suggests that current T2V models are generally capable of grounding high-level semantics from text, yet struggle to faithfully model the consequences of actions on object states over time. Notably, realism also remains challenging, particularly in terms of human evaluation, suggesting that limitations in accurately modeling object state changes are often accompanied by residual visual artifacts, even when aesthetic quality is relatively strong.

\begin{figure}[t]
	\centering
	\includegraphics[width=0.48\textwidth]{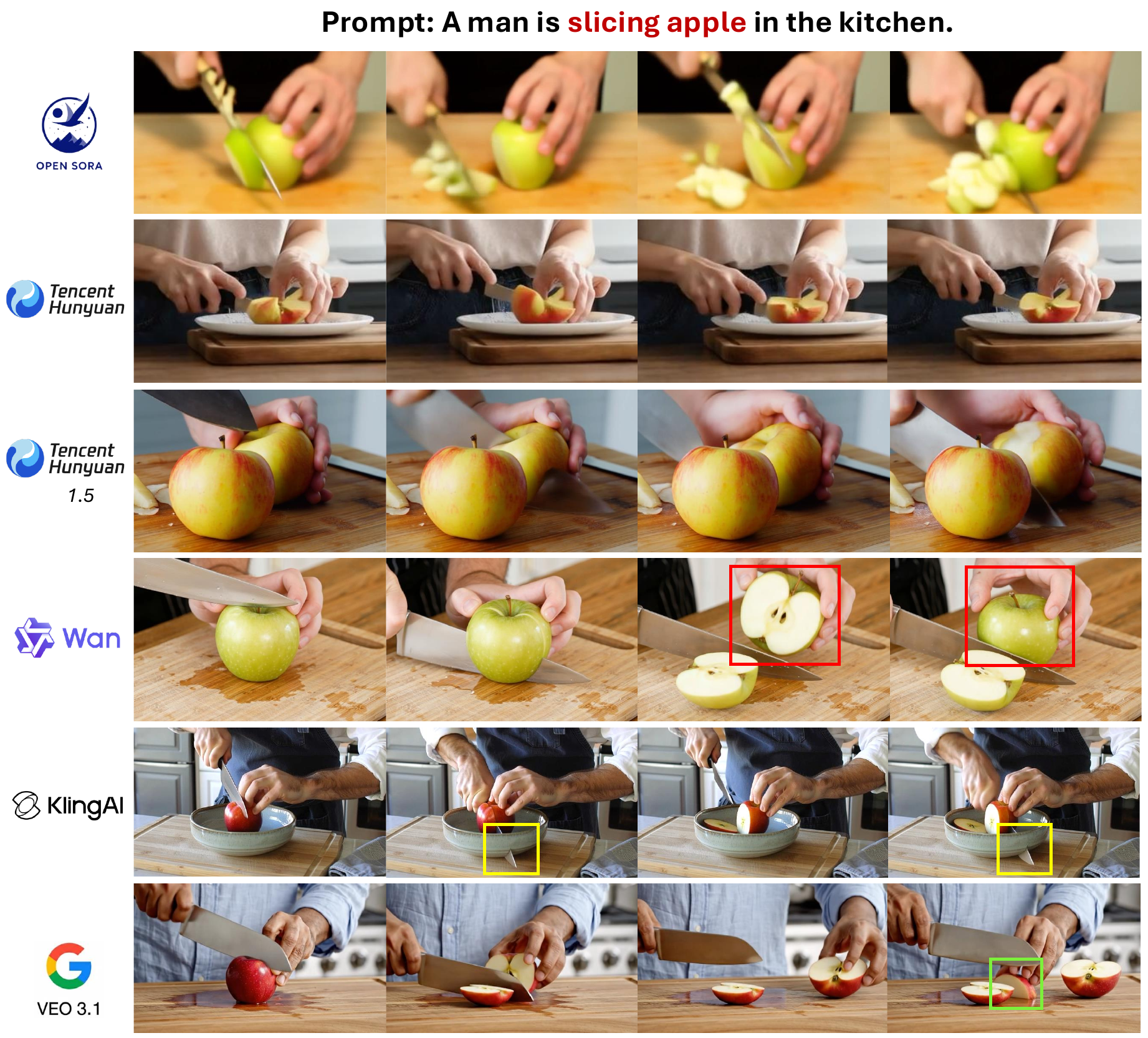}
    \vspace{-0.2cm}
	\caption{Sampled video frames generated by different T2V models. State change consistency or noticeable artifacts are highlighted in boxes.}
	\label{case_model}
\end{figure}

To further illustrate these findings, Figure~\ref{case_model} shows example videos generated by different models for the same object-state-change prompt. 
In the first three models, the object state change is incorrect, where the apple is not sliced into pieces. Although videos generated by Wan-2.2, Kling-2.5-Turbo, and Veo-3.1-Fast successfully exhibit slicing behavior, they still suffer from issues in state change consistency or noticeable artifacts. For instance, Wan-2.2 shows a half-sliced apple reverting to a whole state (red box),
Kling-2.5-Turbo produces unreal interactions between the knife and the bowl (yellow box), and Veo-3.1-Fast introduces an additional apple chunk in the final frame (green box). Despite these issues, most models correctly render the subject, object, and scene, reinforcing the conclusion that high-level semantic alignment is substantially easier than accurate and consistent object state change modeling.

\begin{table}[t]
	\centering
	\scriptsize
	{
	\renewcommand{\arraystretch}{1.1}
    \begin{tabular}{p{3.9cm}>{\centering\arraybackslash}p{0.5cm}>{\centering\arraybackslash}p{0.5cm}>{\centering\arraybackslash}p{1.2cm}}
    \hline
    \multicolumn{1}{c}{\multirow{2}{*}{Models}} & \multicolumn{3}{c}{Object State Change Scenario} \\ \cline{2-4} 
    \multicolumn{1}{c}{}& Regular & Novel & Compositional \\ \hline
    \multicolumn{4}{l}{\cellcolor[HTML]{EFEFEF}Open-source models}\\
    Open-Sora-2.0~\citep{opensora2}&0.410 &0.389 &\textbf{0.416}\\
    HunyuanVideo~\citep{kong2024hunyuanvideo}&\textbf{0.472}&0.405&0.437 \\
    HunyuanVideo-1.5~\citep{hunyuanvideo2025}&\textbf{0.572}&0.559&0.556 \\
    Wan-2.2~\citep{wan2025}&\textbf{0.635}&0.531&0.594 \\ \hline
    \multicolumn{4}{l}{\cellcolor[HTML]{EFEFEF}Proprietary models}  \\
    Kling-2.5-Turbo~\citep{KlingAI2024}&\textbf{0.744}&0.714&0.699\\
    \mbox{Veo-3.1-Fast~\citep{Veo2025}}&0.797&0.731&\textbf{0.805}\\ \hline
    \end{tabular}
    }
    \vspace{-0.2cm}
    \caption{Human-evaluated object state change scores of T2V models across regular, novel, and compositional scenarios, averaged over accuracy and consistency.}
    \label{regular_novel_complex}
\end{table}

\subsection{Human–MLLM Correlation Analysis}
We analyze the correlation between human and automatic evaluation results to assess the reliability of MLLM-based evaluation. 
We report the model correlations with human evaluation in terms of Kendall's $\tau$ and Spearman's $\rho$ in Table~\ref{correlation}, and include inter-evaluator agreement among human evaluators as a reference.
Overall, MLLM-based evaluators exhibit substantially higher correlation with human judgments than the text–video similarity model ViCLIP across all evaluation dimensions, highlighting the advantage of multimodal reasoning over similarity-based scoring.
Among all evaluated MLLMs, GPT-5.2 by incorporating the CoT evaluation strategy generally achieves the strongest overall agreement with human evaluation, indicating that explicitly structured reasoning helps the model better identify fine-grained visual cues and state transitions.
Despite these strengths, we also observe noticeably weaker correlations on perceptual-quality metrics (i.e., realism and aesthetics) for GPT-5.2 with CoT, compared to other MLLMs. This gap likely reflects the inherently subjective nature of such judgments and indicates that fully automating perceptual assessment remains challenging.
Besides, although human-MLLM correlations are still lower than human–human agreement, Figure~\ref{bar} shows that the MLLM-based evaluation produces the same overall ranking of T2V systems as human. This suggests that automatic evaluation with MLLMs, while imperfect at the fine-grained scoring, is nevertheless reliable for assessing overall model performance trends at scale.

\subsection{Category Analysis}
Table~\ref{regular_novel_complex} presents the object state change performance of different T2V models across regular, novel, and compositional scenarios. Regular scenarios mainly achieve the highest scores across all models, as they largely reflect common action–object combinations that are well represented in training data. Novel scenarios exhibit the most severe performance degradation, indicating that current T2V models struggle to generalize state-change reasoning to uncommon but feasible action–object pairs. In contrast, compositional scenarios generally perform better than novel ones but worse than regular ones. This suggests that composing multiple familiar actions in sequence is less challenging than reasoning about unseen combinations, yet still requires maintaining coherent intermediate states over time.

\begin{figure}[t]
	\centering
	\includegraphics[width=0.47\textwidth]{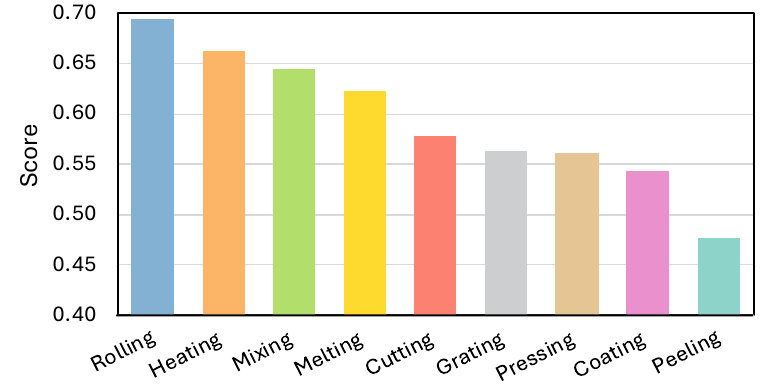}
    \vspace{-0.2cm}
	\caption{Object state change performance across action categories by human evaluation. Scores are averaged over accuracy and consistency.}
	\label{bar_action}
\end{figure}

To further investigate how different actions affect object state change performance, Figure~\ref{bar_action} reports results across action categories. Models achieve higher scores on relatively simple actions with clear and visually salient transformations, such as rolling and heating (e.g., rolling dough or heating root vegetables), where state changes are localized and temporally straightforward. In contrast, performance drops substantially for actions involving complex hand–object interactions or subtle visual transitions, such as peeling, coating, and pressing (e.g., peeling carrot or coating shrimp). These actions require precise manipulation and often involve gradual appearance changes, making both the state change and its visual evidence more difficult for models to capture.

\section{Conclusion}
We have presented OSCBench, a benchmark for evaluating text-to-video generation with a focus on object state change. OSCBench systematically characterizes regular, novel, and compositional state-transition scenarios, covering a broad spectrum of cooking activities. Using this benchmark, we evaluate six representative T2V models using both human user study and MLLM-based automatic assessment, and analyze the correlation between the two methods to assess the reliability of automatic evaluation. Our experiments demonstrate that existing models generally succeed at grounding high-level semantics and producing visually appealing content, but they struggle to accurately and consistently model object state change over time. These limitations persist across regular, novel, and compositional scenarios, and are particularly pronounced for actions involving subtle or complex hand–object interactions. Overall, OSCBench, together with our evaluation framework and empirical analyses, reveals fundamental limitations of existing T2V systems in modeling object state change, and provides a diagnostic foundation for developing more state-aware and robust video generation models in future work.



\section*{Limitations}
While OSCBench provides a focused benchmark for evaluating object state change in text-to-video generation, it has several limitations. First, OSCBench primarily focuses on cooking-related manipulation scenarios, which offer clear and well-defined object state changes but do not fully capture the diversity of interactions found in other domains, such as tool use, household assembly, or outdoor activities. Although cooking covers a wide range of everyday manipulations, extending OSCBench to broader domains would further improve its generality and applicability. Second, our evaluation emphasizes comparative and diagnostic analysis rather than exhaustive human annotation of all generated videos, due to practical cost and scalability constraints. While our sampling strategy ensures balanced coverage across regular, novel, and compositional scenarios, larger-scale human evaluation could reveal additional fine-grained failure modes that are not fully captured in the current setting. We view these limitations as opportunities for future work and hope that OSCBench will serve as a foundation for extending object state change evaluation to broader domains and more comprehensive assessment protocols.

\section*{Ethical Considerations}
Our study employs human evaluation to assess video generation quality and to serve as a reference for validating the reliability of MLLM-based automatic scoring. A representative subset of generated videos were rated by human according to clearly defined criteria. Participants were informed about the study and provided informed consent prior to participation. Since the task involved only the evaluation of model-generated videos, no personal or sensitive information was collected.

The evaluation tasks did not expose participants to harmful or sensitive content. All prompts used in the benchmark were reviewed by the authors to ensure that no unsafe or dangerous material was included. Our work is conducted solely for research purposes and aims to improve the reliability and transparency of multimodal evaluation, rather than to create or promote harmful applications.

\section*{Acknowledgments}
This research/project is supported by the Ministry of Education (MOE), Singapore, under its Academic Research Fund (AcRF) Tier 2 (Proposal ID: T2EP20125-0048). Any opinions, findings and conclusions or recommendations expressed in this material are those of the authors and do not reflect the views of the Ministry of Education, Singapore.

\bibliography{custom}

\clearpage
\appendix
In this appendix, we first present a detailed overview of OSCBench in Section~\ref{OSCBench_detail} and describe the video generation settings in Section~\ref{generation_detail}. Section~\ref{evaluation_detail} provides the evaluation procedure and additional MLLM-based results. Finally, Section~\ref{example_three_scenarios} offers illustrative examples of different object state change scenarios, together with analyses that facilitate a deeper understanding of the benchmark.

\section{OSCBench Details}
\label{OSCBench_detail}
\begin{figure*}[t]
	\centering
	\begin{subfigure}{0.31\linewidth}
		\centering
		\includegraphics[width=\linewidth]{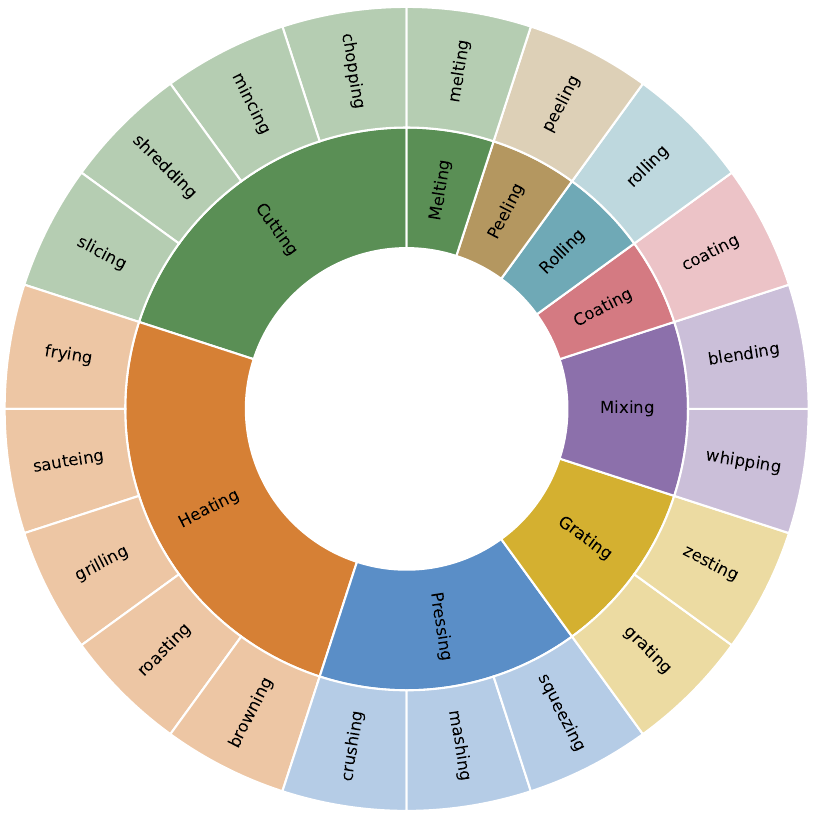}
		\caption{Action taxonomy.}
	\end{subfigure}
    \hspace{3mm}
	\begin{subfigure}{0.31\linewidth}
		\centering
		\includegraphics[width=\linewidth]{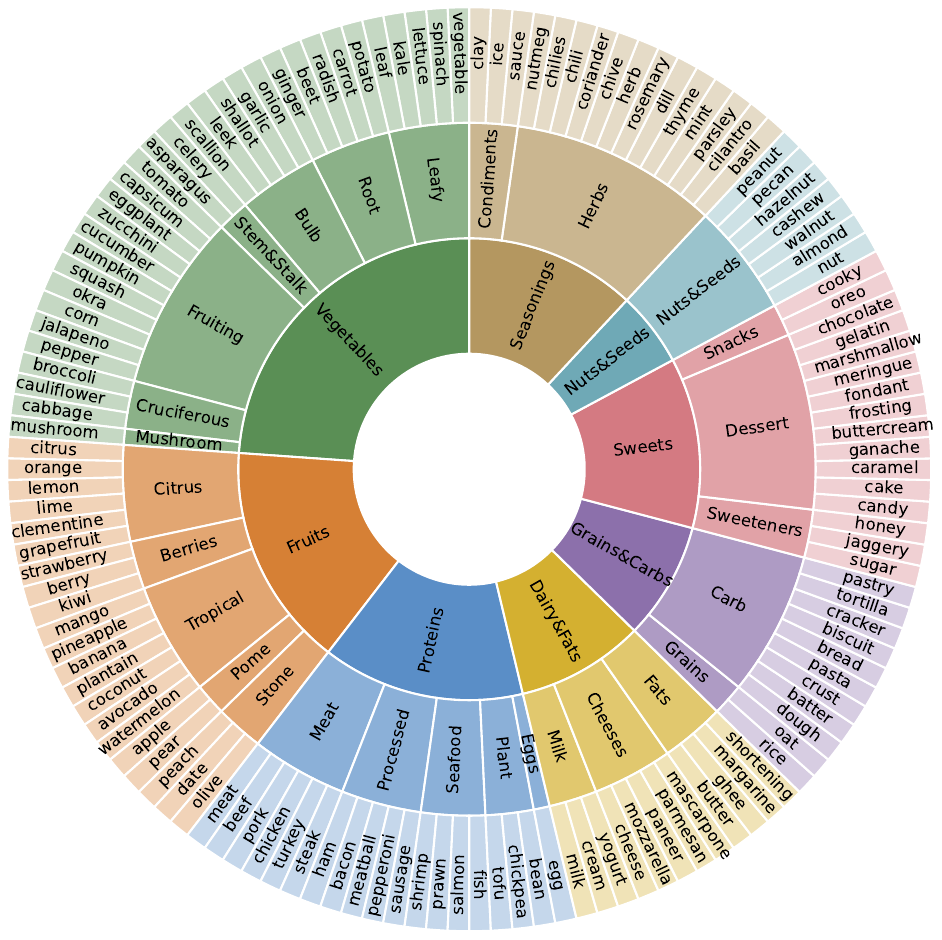}
		\caption{Object taxonomy.}
	\end{subfigure}
    \hspace{3mm}
    \begin{subfigure}{0.31\linewidth}
		\centering
		\includegraphics[width=\linewidth]{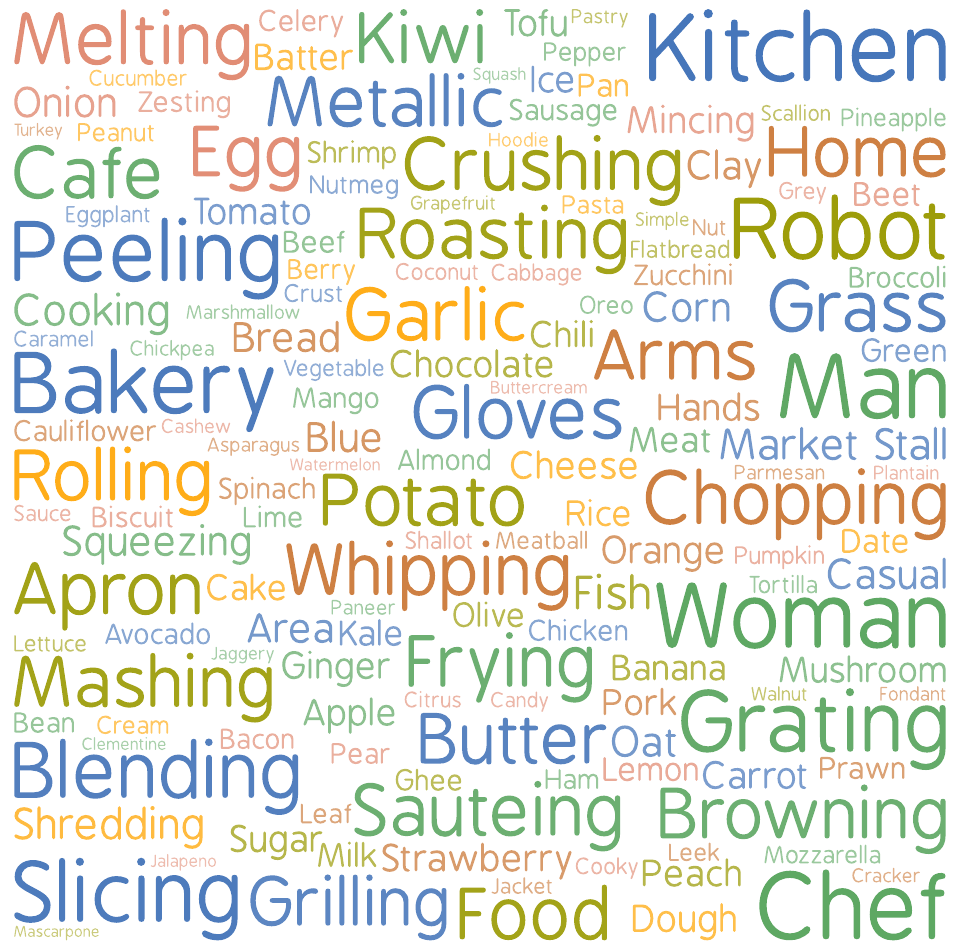}
		\caption{Word cloud of OSCBench.}
	\end{subfigure}
	\caption{Data abstraction results and word cloud in OSCBench.}
	\label{elements}
\end{figure*}

\textbf{Data Abstraction Results}.
In OSCBench contruction, we begin with categorizing the actions and objects in the HowToChange dataset. The taxonomy is constructed through GPT-5.2–assisted grouping, cross-checked with Gemini-3, and subsequently subjected to human-in-the-loop review by human experts. The experts consist of three PhD-level researchers with extensive cooking experience. Based on this process, actions are organized into a two-level hierarchy, whereas objects follow a three-level hierarchical structure. The resulting taxonomy is shown in Figure~\ref{elements} (a) and (b). Building on this taxonomy, we then design a complementary set of object state change scenarios.

\noindent\textbf{Word Distribution in OSCBench}.
We visualize the word distribution of all prompts in OSCBench using a word cloud, as shown in Figure~\ref{elements} (c). This provides an intuitive overview of the dominant concepts and highlights the diversity of objects and actions represented in the benchmark.
We further summarize the number of prompts and evaluation dimensions across different T2V benchmarks in Table~\ref{benchmarks}. As shown, OSCBench explicitly emphasizes object state change, complementing existing benchmarks that primarily focus on semantic adherence or physical plausibility.

\begin{table}[h]
	\centering
	\scriptsize
	{
		\renewcommand{\arraystretch}{1.22}
		\begin{tabular}{p{3.4cm}>{\centering}p{0.65cm}>{\centering}p{0.2cm}>{\centering}p{0.2cm}>{\centering}p{0.2cm}c}
			\hline
			Benchmarks      & \#Prompt & SA& PQ &PC & OSC  \\ \hline
			VBench~\citep{huang2024vbench} &1362 &\checkmark &\checkmark & &  \\
			EvalCrafter~\citep{liu2024evalcrafter} &700 &\checkmark &\checkmark & & \\
			T2V-CompBench~\citep{sun2025t2v} &1400 &\checkmark &&&\\
			VideoPhy~\citep{bansal2024videophy} &688 &\checkmark &  &\checkmark &  \\
			PhyGenBench~\citep{meng2024towards} &160 &\checkmark & &\checkmark & \\
			PhyWorldBench~\citep{gu2025phyworldbench} & 1050 &\checkmark& &\checkmark&  \\
			OSCBench (ours) & 1120 & \checkmark &\checkmark & &\checkmark    \\ \hline
		\end{tabular}
	}
	\caption{Number of prompts and evaluation dimensions in different T2V generation benchmarks. We abbreviate semantic adherence (SA), perceptual quality (PQ), physical commonsense (PC), and object state change (OSC).}
	\label{benchmarks}
\end{table}

\section{Video Generation Setting}
\label{generation_detail}
We generate videos for all prompts in our benchmark for each open-source T2V model. For each proprietary T2V model, we generate videos for the selected 140 prompt used in human evaluation. 
We follow the official and default implementations of T2V models in evaluation. Details of the video generation setting of T2V models, including resolution, total frames, frames per second (FPS), and duration are presented in Table~\ref{generation}. For each generated video, we uniformly sample 20 frames for MLLM-based evaluation. For ViCLIP-based semantic similarity measurement, we uniformly sample 8 frames per video to align with the model architecture.

\begin{table}[h]
	\centering
	\scriptsize
	{
	\renewcommand{\arraystretch}{1.18}
	\begin{tabular}{lcccc}
		\hline
		Models   & Resolution       & Frames & FPS & Duration (s) \\ \hline
		Open-Sora-2.0& 768$\times$768   & 129    & 25  & 5            \\
		HunyuanVideo& 1280$\times$720  & 129    & 25  & 5            \\
		HunyuanVideo-1.5 & 1280$\times$720  & 121    & 24  & 5            \\
		Wan-2.2 & 1280$\times$720  & 81     & 16  & 5            \\ \hline
		Kling-2.5-Turbo& 1920$\times$1080 & 121    & 24  & 5            \\
		Veo-3.1-Fast& 1280$\times$720  & 144    & 24  & 6            \\ \hline
	\end{tabular}
	}
	\caption{Generation settings of T2V models in terms of resolution, total frames, FPS, and duration.}
	\label{generation}
\end{table}

\begin{table*}[t]
	\centering
    \small
	\begin{tabular}{p{0.96\linewidth}}
		\toprule
		Suppose you are an expert in judging and evaluating the quality of AI-generated videos. Such videos may exhibit anomalies such as unnatural object appearance or disappearance, physically implausible state changes, and temporal inconsistencies across frames. They may also contain visual artifacts or unnatural textures. You are given 20 frames evenly sampled from a 5-second AI-generated video.\\

		Video Prompt:\\
		``A chef with a white apron is slicing leek at a street food stand.''\\
		\\
		Your Task:\\
		Analyze these frames chronologically and evaluate the video using the following criteria.\\
		\{Criteria\}\\
		\\
		Instructions:\\
		- Evaluate each criterion INDEPENDENTLY.\\
		- For each criterion, first identify the relevant factual evidence from the frames, then assign a score.\\
		\\
		Output Format:\\
		Return the result strictly in JSON format.\\
		\{\\
			"Subject Alignment": {{"evidence": "...", "score": [1-5]}},\\
			"Object Alignment": {{"evidence": "...", "score": [1-5]}},\\
			"Action Alignment": {{"evidence": "...", "score": [1-5]}},\\
			"OSC Accuracy": {{"evidence": "...", "score": [1-5]}},\\
			"OSC Consistency": {{"evidence": "...", "score": [1-5]}},\\
			"Scene Alignment": {{"evidence": "...", "score": [1-5]}},\\
			"Realism": {{"evidence": "...", "score": [1-5]}},\\
			"Aesthetics": {{"evidence": "...", "score": [1-5]}}\\
		\}
		\\
		\bottomrule
	\end{tabular}
	\caption{Prompt for MLLM to generate the evidence and score for each sampled video. Criteria are the same as those used in the human-evaluation interface.}
	\label{prompt}
\end{table*}

\begin{table*}[t]
	\centering
	\scriptsize
	{
		\renewcommand{\arraystretch}{1.06}
		\begin{tabular}{p{5cm}cccccccc}
			\hline
			\multicolumn{1}{c}{} & \multicolumn{3}{c}{Semantic Adherence}        & \multicolumn{2}{c}{Object State Change} & \multicolumn{1}{c}{\multirow{3}{*}{\shortstack{Scene\\Alignment}}}   & \multicolumn{2}{c}{Perceptual Quality} \\ 
			\cmidrule(lr){2-4} \cmidrule(lr){5-6} \cmidrule(lr){8-9}
			\multicolumn{1}{c}{\multirow{-2}{*}{Model}} & \multicolumn{1}{c}{Subject} & \multicolumn{1}{c}{Object} & \multicolumn{1}{c}{Action} & \multicolumn{1}{c}{Accuracy} & \multicolumn{1}{c}{Consistency} & & \multicolumn{1}{c}{Realism} & \multicolumn{1}{c}{Aesthetics} \\ \hline
			\multicolumn{9}{l}{\cellcolor[HTML]{EFEFEF}\textit{Open-source models}}                 \\
			Open-Sora-2.0~\citep{opensora2}   & 0.978& 0.932 & 0.718 & 0.656 & 0.722 & 0.980 & 0.796& 0.816\\
			HunyuanVideo~\citep{kong2024hunyuanvideo}   & 0.966 & 0.954 & 0.682& 0.648 & 0.752  & 0.988  & 0.846& 0.858 \\
			HunyuanVideo-1.5~\citep{hunyuanvideo2025}& 0.986&\textbf{0.958}& \textbf{0.780}& \textbf{0.756} & 0.808& 0.990& 0.864& 0.864 \\
			Wan-2.2~\citep{wan2025} &\textbf{0.992}& 0.952 & 0.738 & 0.712 & \textbf{0.820} & \textbf{0.992} &\textbf{0.882} & \textbf{0.910}\\ \hline
			\multicolumn{9}{l}{\cellcolor[HTML]{EFEFEF}\textit{Proprietary models}} \\
			Kling-2.5-Turbo~\citep{KlingAI2024}& \textbf{0.998}& 0.972& 0.934& 0.882& 0.906& \textbf{0.998}& 0.950& 0.952 \\
			Veo-3.1-Fast~\citep{Veo2025} & 0.996& \textbf{0.988}& \textbf{0.970}& \textbf{0.954} & \textbf{0.968}& 0.996& \textbf{0.994}& \textbf{0.978}\\ \hline
		\end{tabular}
	}
	\caption{Qwan3-VL-30B-based evaluation results of different T2V models across multiple evaluation dimensions.}
	\label{qwen_table}
\end{table*}

\begin{table*}[t]
	\centering
	\scriptsize
	{
		\renewcommand{\arraystretch}{1.06}
		\begin{tabular}{p{5cm}cccccccc}
			\hline
			\multicolumn{1}{c}{} & \multicolumn{3}{c}{Semantic Adherence}        & \multicolumn{2}{c}{Object State Change} & \multicolumn{1}{c}{\multirow{3}{*}{\shortstack{Scene\\Alignment}}}   & \multicolumn{2}{c}{Perceptual Quality} \\ 
			\cmidrule(lr){2-4} \cmidrule(lr){5-6} \cmidrule(lr){8-9}
			\multicolumn{1}{c}{\multirow{-2}{*}{Model}} & \multicolumn{1}{c}{Subject} & \multicolumn{1}{c}{Object} & \multicolumn{1}{c}{Action} & \multicolumn{1}{c}{Accuracy} & \multicolumn{1}{c}{Consistency} & & \multicolumn{1}{c}{Realism} & \multicolumn{1}{c}{Aesthetics} \\ \hline
			\multicolumn{9}{l}{\cellcolor[HTML]{EFEFEF}\textit{Open-source models}}                 \\
			Open-Sora-2.0~\citep{opensora2}   & 0.918& 0.838 & 0.712 & 0.794 & 0.814 & 0.930 & 0.722& 0.774\\
			HunyuanVideo~\citep{kong2024hunyuanvideo}   & 0.960 & 0.942 & 0.792& 0.718 & 0.956  & 0.996  & 0.908& 0.900 \\
			HunyuanVideo-1.5~\citep{hunyuanvideo2025}& 0.934&0.914&0.794& 0.582 & 0.788& 0.962& 0.834& 0.788 \\
			Wan-2.2~\citep{wan2025} &\textbf{0.946}& \textbf{0.944} & \textbf{0.750} & \textbf{0.744} & \textbf{0.970} & \textbf{0.966} &\textbf{0.930} & \textbf{0.946}\\ \hline
			\multicolumn{9}{l}{\cellcolor[HTML]{EFEFEF}\textit{Proprietary models}} \\
			Kling-2.5-Turbo~\citep{KlingAI2024}& \textbf{0.996}& 0.982& 0.976& 0.906& \textbf{0.998}& 0.996& \textbf{0.980}& \textbf{1.000} \\
			Veo-3.1-Fast~\citep{Veo2025} & 0.992& \textbf{0.994}& \textbf{0.994} & \textbf{0.966}& \textbf{0.998}& \textbf{0.998}&0.958& 0.984\\ \hline
		\end{tabular}
	}
	\caption{GPT-5-mini-based evaluation results of different T2V models across multiple evaluation dimensions.}
	\label{5mini_table}
\end{table*}

\begin{figure*}[t]
	\centering
	\includegraphics[width=\textwidth]{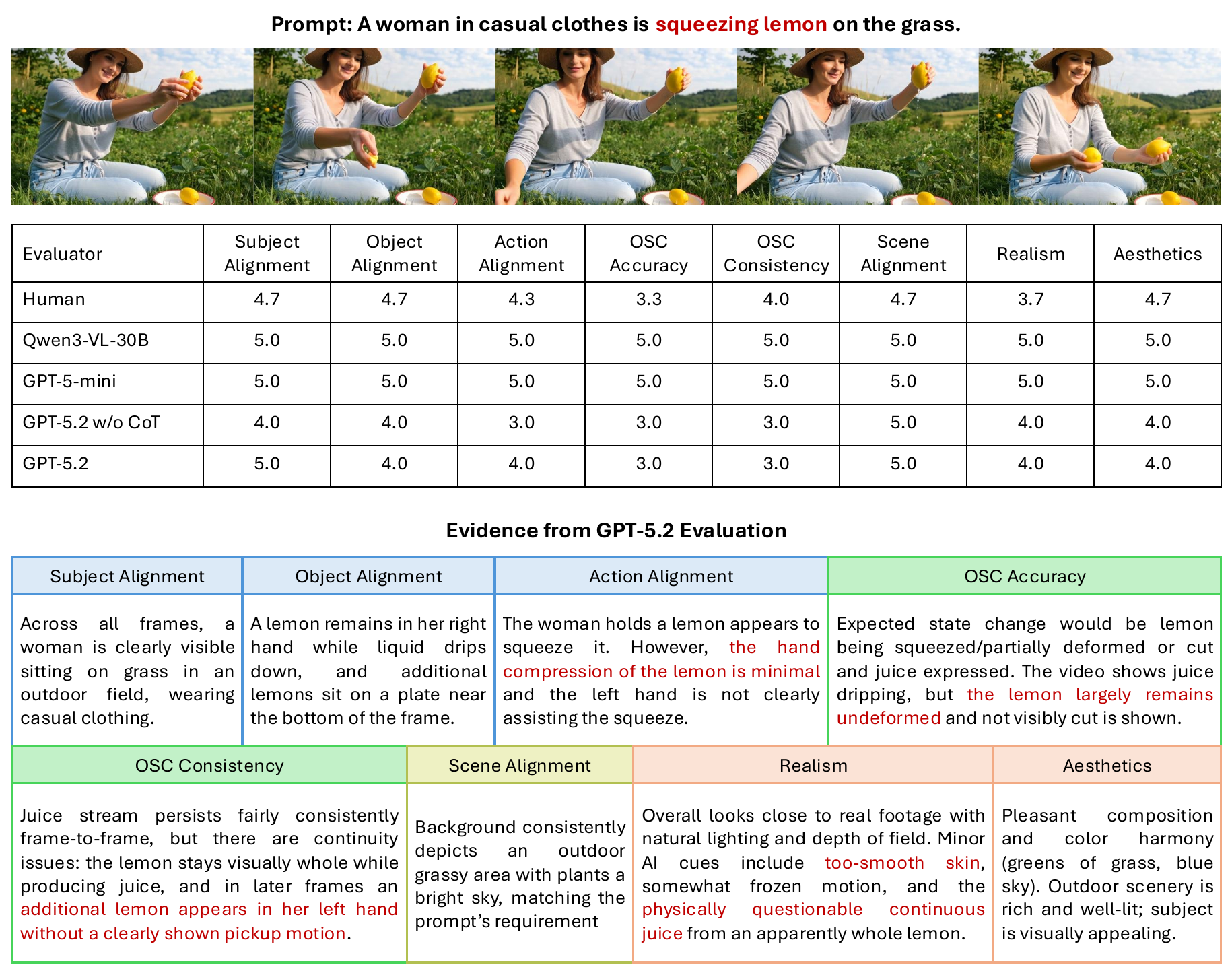}
	\caption{Example of human and MLLM evaluation on a generated video. Original human evaluation scores averaged over three evaluators are provided for reference. The evidence is generated by GPT-5.2 when scoring with CoT.}
	\label{mllm_example}
\end{figure*}

\section{Evaluation Details}
\label{evaluation_detail}
In this section, we provide additional details about our evaluation protocol. We first describe the scoring criteria and then report the results obtained from the MLLM-based evaluation, followed by an illustrative example of how the MLLMs reason and assign scores.

\begin{figure*}[t]
	\centering
	\includegraphics[width=0.96\textwidth]{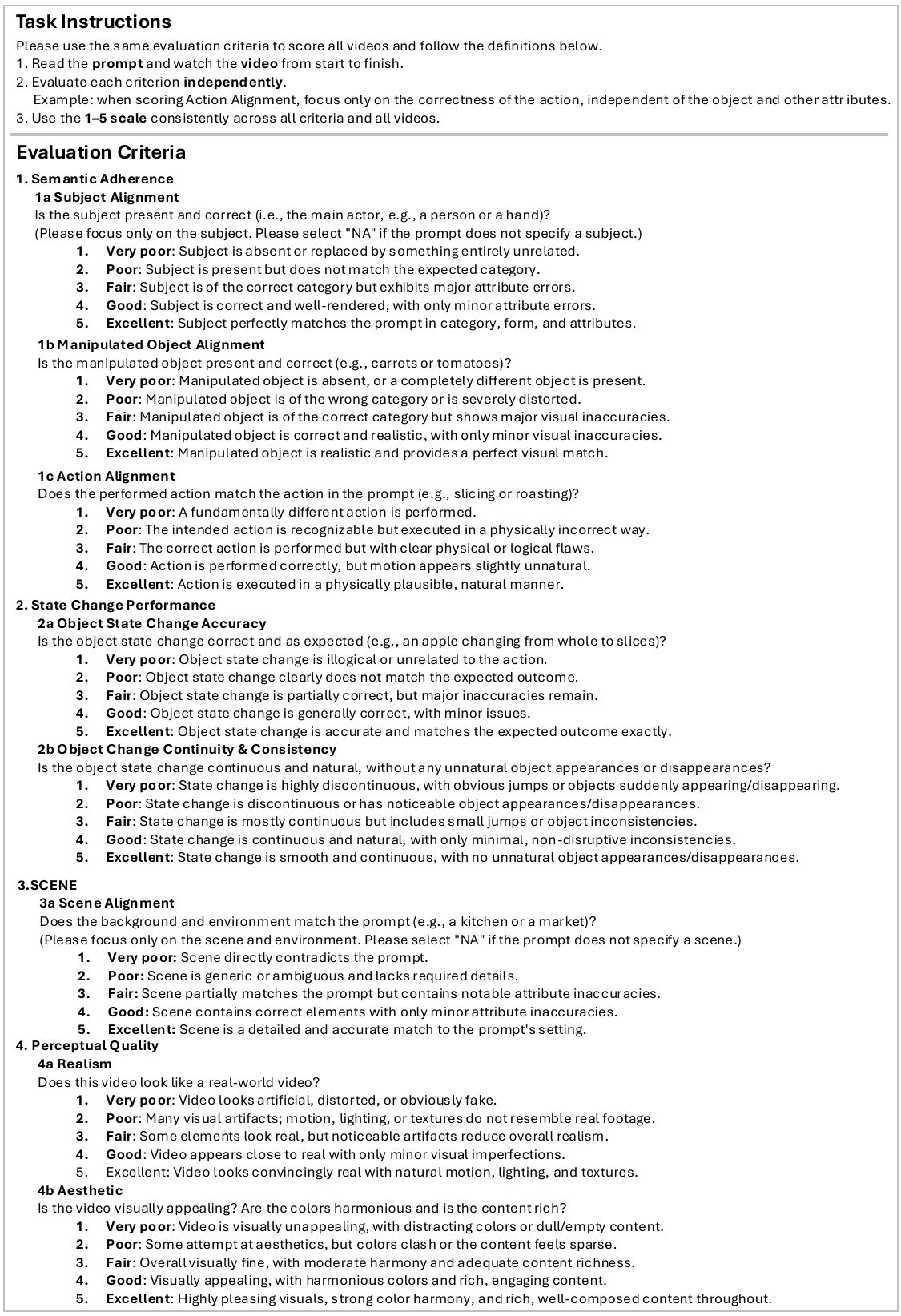}
	\caption{Task instructions and evaluation criteria in the human evaluation interface.}
	\label{criteria}
\end{figure*}

\textbf{Scoring Criteria}.
We adopt a hybrid evaluation protocol that combines human user study with automated MLLM-based evaluation. Across these two evaluation modes, we design a comprehensive set of evaluation dimensions covering semantic adherence, OSC performance, scene alignment, and perceptual quality. For each dimension, we provide detailed scoring criteria for both human evaluator and MLLM-based evaluation. 
The instructions and scoring rubrics in the human user study interface are shown in Figure~\ref{criteria}, according to which human evaluators are asked to rate each video on a scale from 1 to 5. For MLLM-based evaluation, the prompts we use are presented in Table~\ref{prompt}.

\textbf{MLLM-based Evaluation Results}.
We report the additional evaluation results of Qwen3-VL-30B and GPT-5-mini in Table~\ref{qwen_table} and Table~\ref{5mini_table}, respectively. Although their correlations with human evaluation are not particularly high, the OSC accuracy in both models is lower than their scores on subject and object semantic adherence. This suggests that Qwen3-VL-30B and GPT-5-mini can capture part of the difficulty associated with object state change.
Furthermore, the rankings of all T2V models produced by the MLLMs are fully consistent with human judgments, as shown in Figure~\ref{bar}. This result indicates that automatic evaluation is reliable for large-scale model comparison and benchmarking, even though fine-grained per-instance scoring remains imperfect.

To further illustrate how MLLMs interpret the videos, Figure~\ref{mllm_example} presents an example of MLLM evaluation on a generated video. Although the video appears visually appealing, human evaluators identify an OSC error: juice is dripping from the lemon, yet the lemon itself shows no visible squeezing or deformation. For this video, we observe that Qwen3-VL-30B and GPT-5-mini assign perfect scores, indicating that they fail to detect the fine-grained issues in the lemon's state change. GPT-5.2, in contrast, is able to detect the OSC error and provides reasonable supporting evidence, noting that ``the lemon largely remains undeformed''. This suggests that more advanced MLLMs can to handle state-change reasoning and identify inconsistencies between visual appearance and expected physical outcomes.
Although GPT-5.2 w/o CoT also assigns a relatively low score to OSC accuracy, it still gives action alignment and OSC accuracy the same score, which shows weaker consistency with human judgments. From the human perspective, the error in action alignment is minor, whereas the error in OSC is a major one. This indicates that using CoT to plan a reasoning route, in which the model first follows the grading guidelines to collect explicit evidence and then assigns scores, encourages more careful evaluation and results in scores that better align with human judgments.

\section{Examples of Different OSC Scenarios}
\label{example_three_scenarios}
\begin{figure*}[t]
	\centering
	\includegraphics[width=0.96\textwidth]{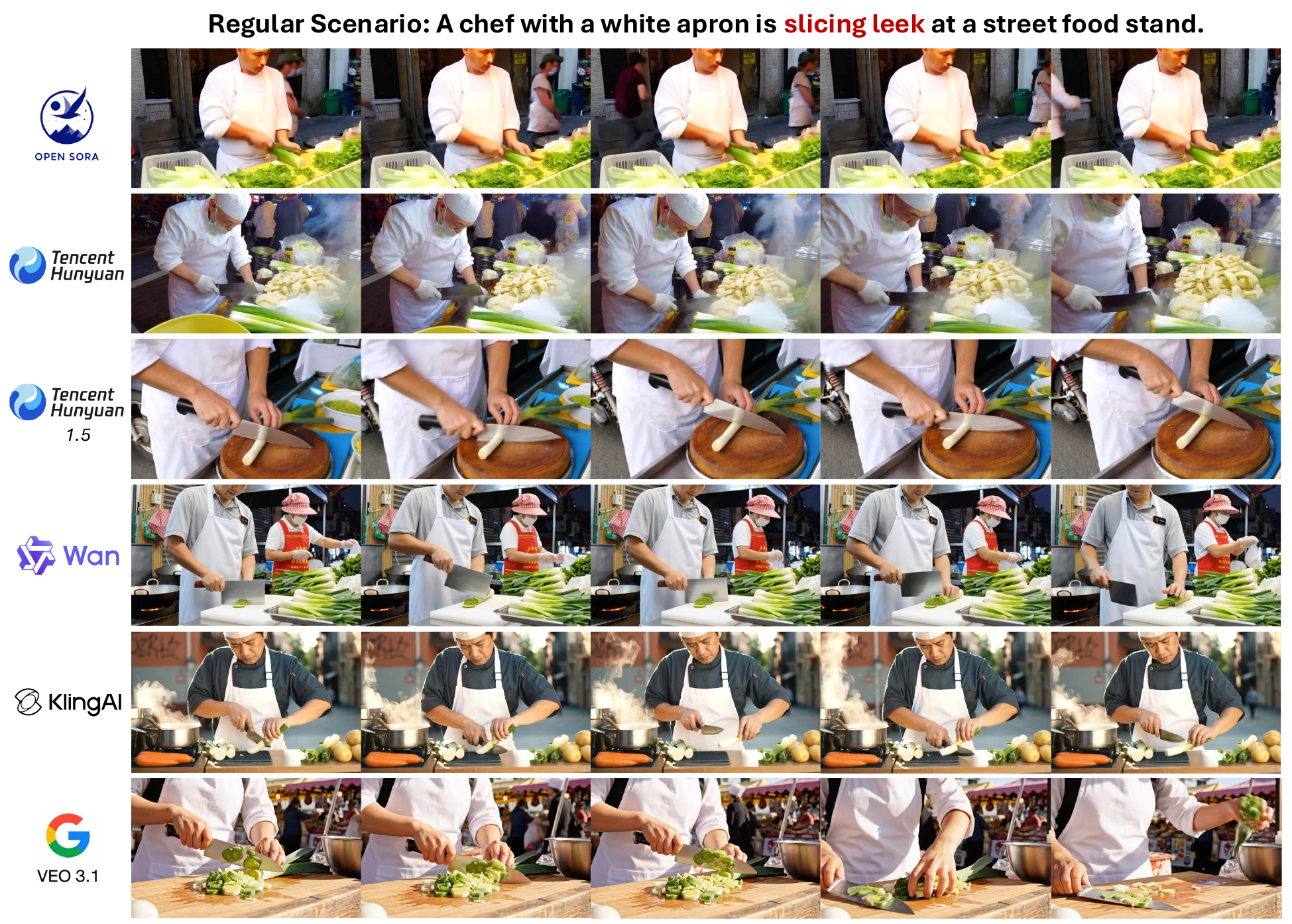}
	\caption{Sampled videos of different models in regular OSC scenario.}
	\label{case_regular}
\end{figure*}

\begin{figure*}[t]
	\centering
	\includegraphics[width=0.96\textwidth]{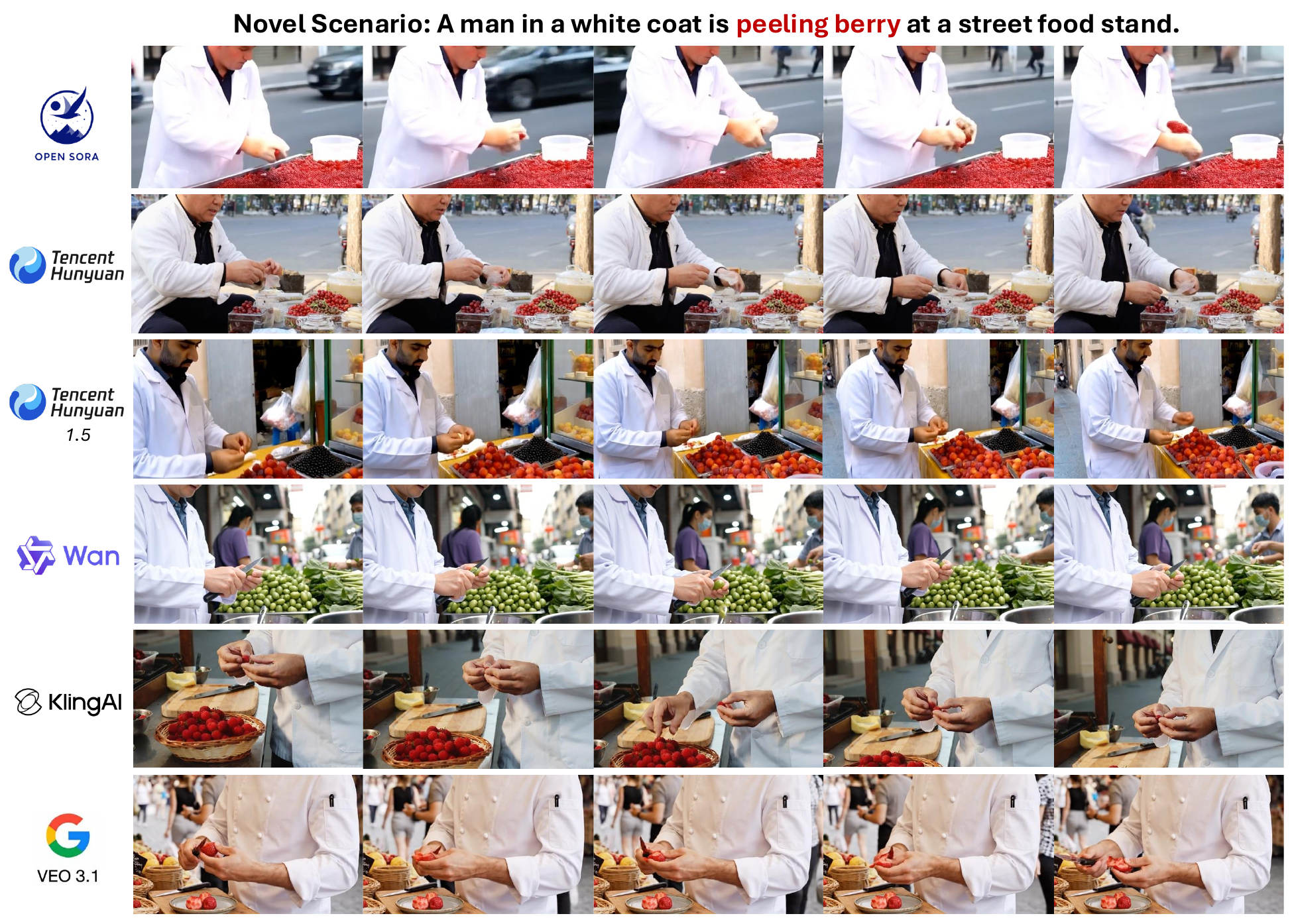}
	\caption{Sampled videos of different models in novel OSC scenario.}
	\label{case_novel}
\end{figure*}

\begin{figure*}[t]
	\centering
	\includegraphics[width=0.96\textwidth]{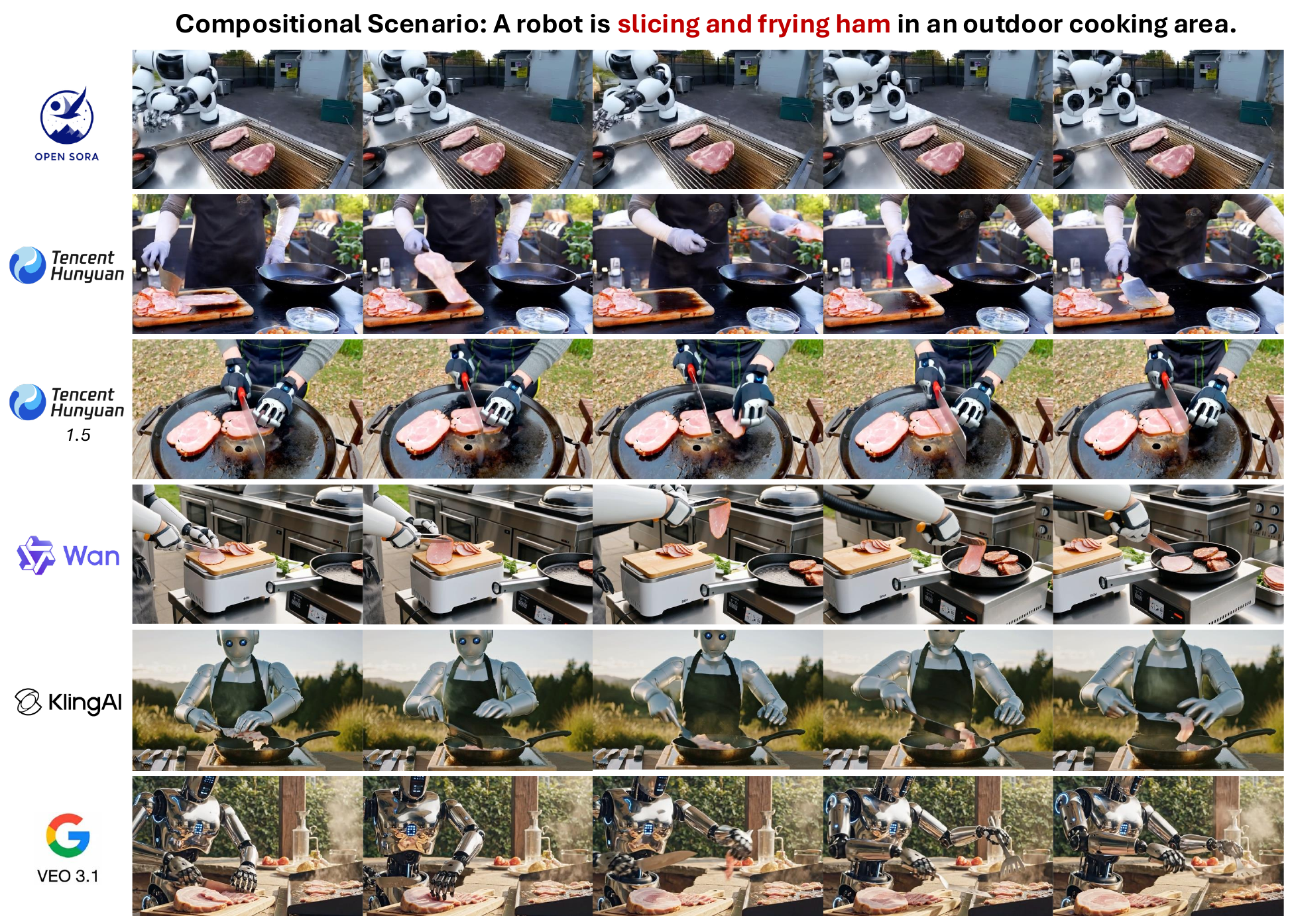}
	\caption{Sampled videos of different models in compositional OSC scenario.}
	\label{case_complex}
\end{figure*}

\begin{figure*}[t]
	\centering
	\includegraphics[width=0.96\textwidth]{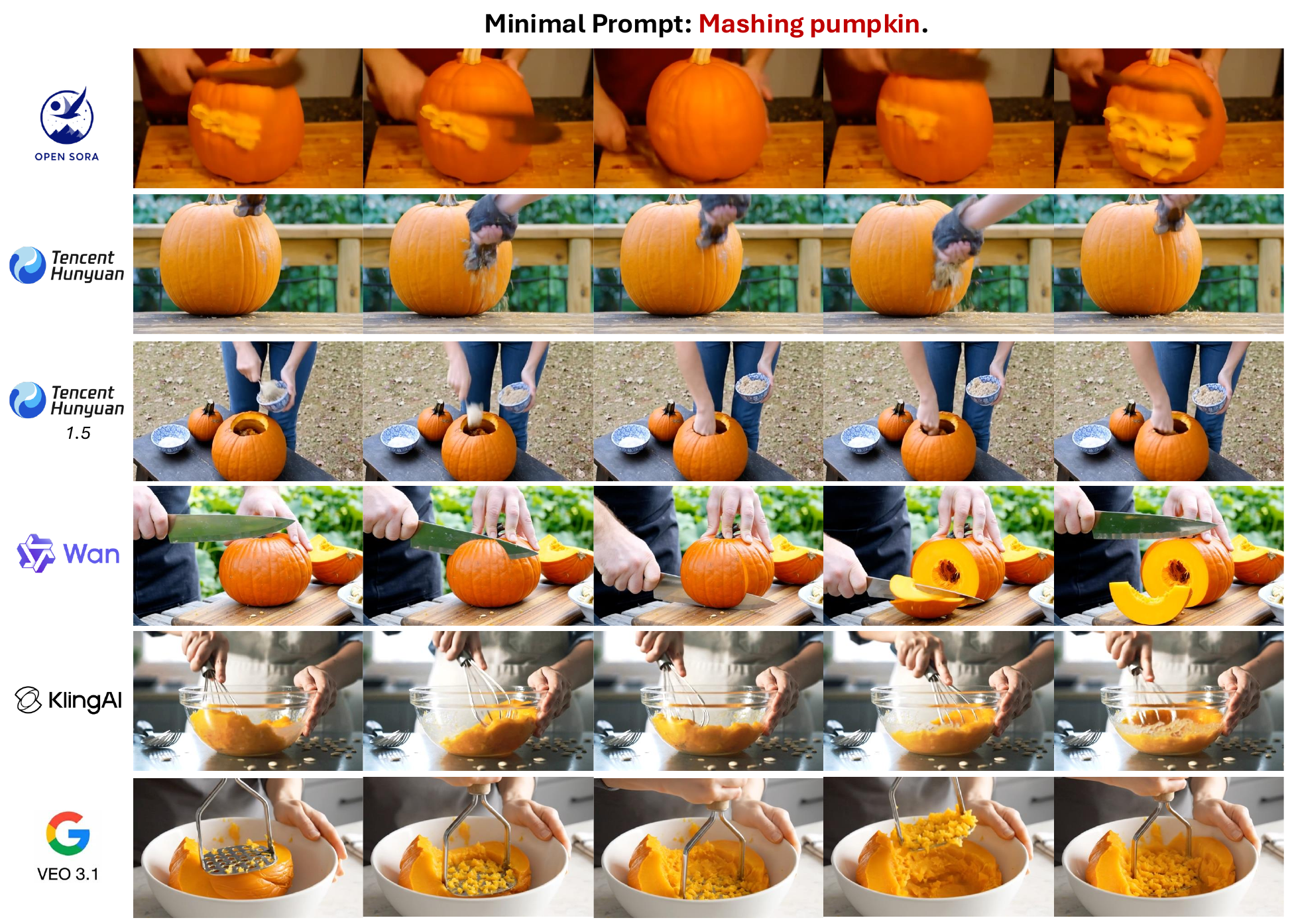}
	\caption{Sampled videos of different models with minimal OSC prompt.}
	\label{case_minimal}
\end{figure*}

To provide a more intuitive view of T2V performance on object state change in OSCBench, we present examples of generated videos in regular, novel, and compositional OSC scenarios in Figures~\ref{case_regular}, Figure~\ref{case_novel}, and Figure~\ref{case_complex}, respectively. We also present examples of generated videos from minimal prompts of the form <action><object> in Figure~\ref{case_minimal}.

In the regular OSC scenario shown in Figure~\ref{case_regular}, all models generate videos in which the subject (chef), action (slicing), object (leek), and scene (street food stand) are rendered well. However, clear errors emerge in the object state change. For example, in videos generated by Open-Sora-2.0, HunyuanVideo, HunyuanVideo-1.5, and Kling-2.5-Turbo, the leek is not actually sliced into pieces. Although videos produced by Wan-2.2 and Veo-3.1-Fast exhibit correct object state changes, the state change consistency in the later frames remains limited.

In the novel OSC scenario shown in Figure~\ref{case_novel}, some models can roughly understand the peeling action, but noticeable issues remain. The hand details in Open-Sora-2.0 are blurred. In HunyuanVideo, HunyuanVideo-1.5, and Wan-2.2, the object being peeled is incorrect. In particular, Wan-2.2 generates olives, which are more commonly associated with the peeling action. This suggests that these models are strongly influenced by memorized training patterns when generating object state changes for uncommon action–object combinations, indicating an incomplete understanding of the intended action. Besides, Kling-2.5-Turbo exhibits state change consistency issues, where two berries gradually collapse into one. Veo-3.1-Fast produces an OSC that is close to correct, but the object still shows jitter and artifacts in the last frame. Overall, the novel scenario remains challenging for current T2V models.

In the compositional OSC scenario shown in Figure~\ref{case_complex}, Open-Sora-2.0, HunyuanVideo, Wan-2.2, and Kling-2.5-Turbo execute only one of the required actions. For example, Open-Sora-2.0, Wan-2.2, and Kling-2.5-Turbo perform only the frying action. HunyuanVideo-1.5 attempts to handle both actions by cutting the ham while it is being fried, but the action is not the intended slicing action. Although Veo-3.1-Fast successfully completes the compositional actions, the consistency of the object states is poor. For example, a spatula suddenly appears in frame 4, and the ham disappears in frame 5, revealing noticeable artificial artifacts. These results suggest that compositional OSC remains challenging for current T2V models.

We also provide an example of generated videos with minimal prompts, <action><object>, in Figure~\ref{case_minimal}. We observe that models can often generate object state changes even under such minimal prompts, suggesting that contextual cues such as subjects or scene descriptions are not strictly required to trigger state-change behavior. In addition, minimal prompts encourage models to focus more on the specified action and object. However, with respect to producing the correct object state change, only Veo-3.1-Fast generates the intended mashing action, while the outputs produced by all other models fail to follow the prompt accurately. This observation highlights the difficulty of accurate OSC generation for most existing T2V models.

\end{document}